\pdfoutput=1
\documentclass[11pt]{article}

\usepackage[final]{neurips_2022}

\usepackage{times}
\usepackage{latexsym}
\usepackage{graphicx}
\graphicspath{ {./images/} }

\usepackage[T1]{fontenc}
\usepackage[utf8]{inputenc}

\usepackage{microtype}

\usepackage{inconsolata}

\usepackage{hyperref}       % hyperlinks
\usepackage{url}            % simple URL typesetting
\usepackage{booktabs}       % professional-quality tables
\usepackage{amsfonts}       % blackboard math symbols
\usepackage{nicefrac}       % compact symbols for 1/2, etc.
\usepackage{microtype}      % microtypography
\usepackage{xcolor}         % colors

\title{Evaluation of Synthetic Datasets for Conversational Recommender Systems}

\author{Harsh Lara \\
  Google Research \\
  \texttt{harshlara@google.com} \\\And
  Manoj Tiwari \\
  Google Research \\
  \texttt{mjtiwari@google.com} \\}

\begin{document}
\maketitle
\begin{abstract}
For researchers leveraging Large-Language Models (LLMs) in the generation of training datasets, especially for conversational recommender systems - the absence of robust evaluation frameworks has been a long-standing problem \citet{Peng_2017}. The efficiency brought about by LLMs in the data generation phase is impeded during the process of evaluation of the generated data, since it generally requires human-raters to ensure that the data generated is of high quality and has sufficient diversity. Since the quality of training data is critical for downstream applications, it is important to develop metrics that evaluate the quality holistically and identify biases. In this paper, we present a framework that takes a multi-faceted approach towards evaluating datasets produced by generative models and discuss the advantages and limitations of various evaluation methods.
\end{abstract}

\section{Introduction}

Conversational recommender systems consist of several machine learning models working together under the hood to perform complex tasks while providing users with a seamless and uncomplicated natural language interface. Even the most primitive conversational recommenders consist of at least four different components: a NLU unit, a NLG unit, a dialogue policy unit and a retrieval and ranking system. Recently, the performance of all of these components have shown promising results using deep neural networks, which require an abundance of training data to train on.

The advent of pre-trained language models (BERT, GPT-3, etc.) have made it possible for natural-language applications to achieve admirable performance using zero-shot or few-shot learning. However, the bulk of complex tasks (e.g. intent prediction, summarization, providing item recommendations etc.) still require that these pre-trained models be finetuned for meeting more composite objectives.

Data collection for fine-tuning tasks often turns out to be the bottleneck that is the differentiator between successful and unsuccessful machine learning applications. Data generation using LLMs (e.g. GPT-3, LaMDA, PALM) \citet{DBLP:journals/corr/abs-2005-14165, DBLP:journals/corr/abs-2201-08239, https://doi.org/10.48550/arxiv.2204.02311} is a promising approach towards treating the data scarcity problem. If we can leverage LLMs using priming along with pre-programmed dialogue policies to generate synthetic data that can replicate human interactions with conversational agents, it can enable a plethora of fine-tuned applications.

Evaluating the quality of synthetic data generated with the help of LLMs is a challenging task. With the help of LLMs, it is possible to generate thousands of examples of interactions between the conversational recommender and a simulated user, however - it is difficult to ensure that the synthesized data is of good quality. Researchers generally hold synthetically generated data at a higher bar for quality as compared to human-generated data. The reason for this is simple - it is easy for biases to creep into data generation when using LLMs instead of humans. Additionally, it is non-trivial to ensure that synthesized data is diverse and representative of the gamut of human interactions with a conversational agent.

\begin{figure*}
\includegraphics[width=\linewidth]{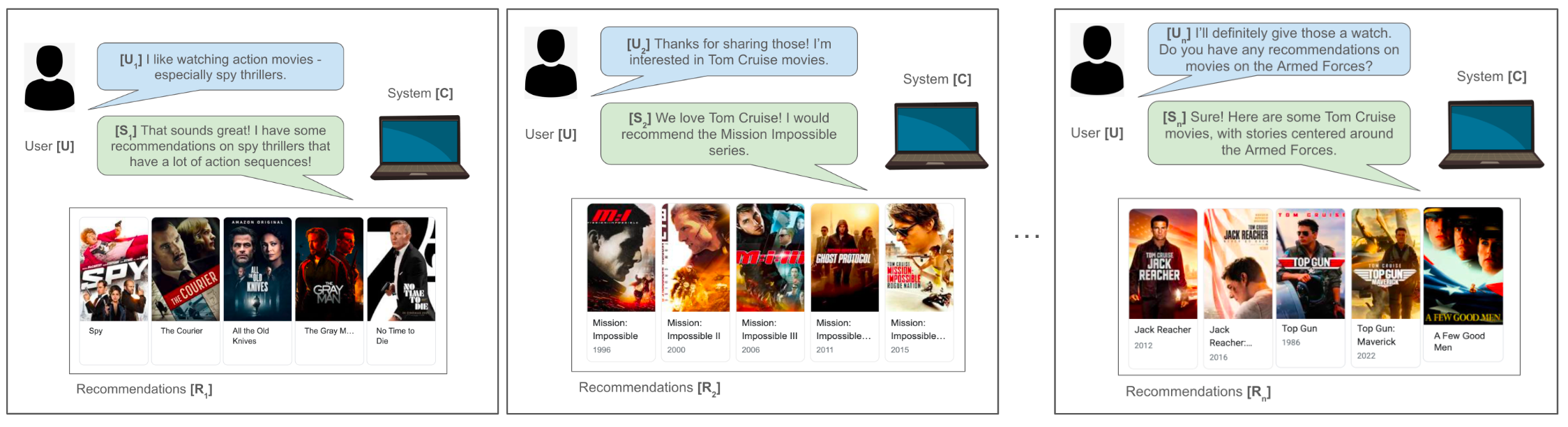}
\caption{Example of a conversation between a user and a conversational movie-recommender system}
\label{fig:conv_rec}
\centering
\end{figure*}
\section{Conversational Recommender Systems}

Conversational recommender systems (or conversational recommenders) are conversational agents which are tasked with engaging in task-oriented, multi-turn dialogue with users in order to provide useful recommendations to them \citet{DBLP:journals/corr/abs-2004-00646}. During such dialogue, the system would elicit user preferences, provide explanations for recommended items, and process user's feedback on the recommendations.

Conversational recommenders are expected to perform better than non-conversational recommender systems, because of their interactive nature and their ability to enable high-bandwidth communication with the user by means of natural language dialogue \citet{RICH1979329}.

Formally, a conversational recommender can be defined as a conversational agent (\emph{C}) that produces a combination of natural language response (\emph{S\textsubscript{i}}) coupled with a slate of recommended items (\emph{R\textsubscript{i}}) to the user, after the \emph{i\textsuperscript{th}} user utterance (\emph{U\textsubscript{i}}). A conversation session consists of a sequence of alternating user and system turns, along with recommendations accompanying each system turn as shown in Figure ~\ref{fig:conv_rec}.

\section{Datasets for Conversational Recommenders}

The objective of conversational recommender systems is two-fold:

\begin{enumerate}
  \item Efficiently recommend items to the user that they would find useful and interesting, given their preferences at that time.
  \item Create exploration opportunities for the user, so that they are able to discover items that they wouldn’t have otherwise found without the assistance of a conversational recommender system.
\end{enumerate}

Therefore, while developing fine-tuning datasets, it is critical that the data exhibits good quality with respect to both of these goals. If we were to optimize for only efficiency of the recommender, we would risk ending up with a dataset that is degenerate - for example, a dataset in which the recommender system directly suggests the top-ranked item which matches the user’s interests in the first turn of the conversation, without creating any exploration opportunities for the user. If we were to optimize only for exploration opportunities for the user, we would end up with a dataset consisting of long-winding conversations that don’t necessarily exhibit satisfactory efficiency by the conversational recommender.

\section{Evaluating Conversational Recommenders Datasets}

Due to the reasons discussed previously, datasets generated for fine-tuning conversational recommenders should be evaluated by using a multi-dimensional approach.

We propose two main criteria for the evaluation of dataset quality:

\begin{enumerate}
  \item \emph{Realism}: The ability of the dataset to imitate the conversational ability of real humans.
  \item \emph{Diversity}: The ability of fine tuning datasets to exhibit the gamut of interactions real humans would have had with the recommender system.
\end{enumerate}

\subsection{Realism}
When simulating user interactions with the system, having the conversation sessions closely mimic real human interactions with the agent is a desirable property. Intuitively, the more closely we model real human conversations in our training data generation, the more likely we are to perform better at downstream tasks such as intent prediction and item reranking.

The natural language utterances from the user enable the conversational recommender to build a useful model of the user's preferences and provide valuable recommendations. Similarly, the natural language output from the conversational recommender enable the user to have a better understanding of the system's suggestions and stimulate a rich back and forth between both entities. Therefore, during the data generation phase, it is imperative to produce a high quality of natural language dialogue during the sessions.

Ultimately the goal of the conversational recommender is to model the user's preferences better and provide useful recommendations to users while enabling a high bandwidth interaction between the user and the system. With high quality dialogue, we expect recommendations produced at the later turns of a session to be much better than the recommendations produced during the initial turns of the conversation.
\subsubsection{Using human raters}
Realism can be measured by having crowd-workers evaluate a set of randomly sampled conversations from a data generation run.

For example, if we have a version of a conversation simulator (say, Simulator K), we would use that to generate \emph{N} (say, N=50) conversations. All of these 50 conversations can be sent to crowd-workers with the following instructions:

“Here are 50 user conversations with a conversational recommender system”.

Some conversations in this set are human-generated, whereas others are simulator-generated. For each conversation, identify whether it is human-generated or simulator-generated.”

Therefore, we would have the following ratio (let's call it Realism Score) computed for the evaluation set that is shared with crowd-workers.

\[Realism Score_{human} = \frac{h_{human}}{N}\]

Where \emph{h} is the number of conversations identified as human-generated and \emph{N} is the total number of conversations shared with human raters.

\subsubsection{Using a machine learned classifier}
Another way of evaluating realism is by training a discriminator to classify conversations into one of two categories: human-generated or simulated.

A low-cost means of training such a discriminator, is to use a human-rated batch of conversations as the ground-truth. As discussed in the previous section, we would share a batch simulated conversations and have crowd-workers provide binary labels for each of them - \emph{simulator-generated} or \emph{human-generated}. the labeled dataset produced in this manner would be used as the training data for the discriminator.

We can run the discriminator on a batch of simulated conversations to obtain a Realism Score derived along the same idea we have used with human-raters.

\[Realism Score_{inferred} = \frac{h_{predicted}}{N}\]

Where \emph{h\textsubscript{predicted}} is the number of conversations predicted to be human-generated and \emph{N} is the total number of conversations used during inference.

\subsection{Diversity}

\begin{table*}
\centering
\begin{tabular}{clll}
\hline
\textbf{Simulator Version} & \textbf{H\textsubscript{Sentiment}} & \textbf{H\textsubscript{Topic}} & \textbf{Entropy Score}\\
\hline
A & 0.76 & 0.58 & 0.67 \\
B & \textbf{0.82} & 0.67 & 0.745 \\
C & 0.66 & \textbf{0.93} & \textbf{0.795} \\\hline
\end{tabular}
\caption{Examples of entropy scores computed for 3 versions of a conversation simulator.}
\label{tab:entropy1}
\end{table*}

In the context of synthetic data generation, we define diversity as the measure of how balanced the distribution of data points (conversations) is with respect to the vocabulary over our chosen dimensions (e.g. sentiments, topics, intents, etc.). As noted by \citet{DBLP:journals/corr/abs-1904-02792} human evaluation, which is often viewed as
the gold standard evaluation, captures quality perfectly but fails to capture diversity

We propose the measurement of diversity by evaluating the distribution of data-points (in our case, dialogue sessions) across different pre-defined dimensions. For each dimension, we would need to specify a vocabulary (classes that any data-point can be classified into) and a corresponding classifier (a system that can classify each data-point into one of the previously defined classes).

For example, a framework to evaluate conversational datasets could include the following classifiers:

\begin{enumerate}
  \item Sentiment Classifier
  \begin{enumerate}
    \item Vocabulary: Negative, Positive.
    \item Goal: Classify each session into one of the predefined sentiment categories, based on the perceived user sentiment.
  \end{enumerate}
  \item Topic Classifier
  \begin{enumerate}
    \item Vocabulary: Automobiles, Cooking, Travel, Technology, Fashion, History.
    \item Goal: Classify each simulated session into a topic class, based on the main focus of the discussion.
  \end{enumerate}
\end{enumerate}

For any data generation job, we can produce an average entropy score, by using the individual entropies produced by these classifiers.

For each dimension, we would estimate the entropy of the source (data generator) by using the standard formula for Shannon entropy \citet{10.1145/584091.584093}:

\[H(X) = \sum_{i=1}^{n}p(x_{i})log_{b}\frac{1}{p(x_{i})}\]

We would compute a final \emph{Entropy Score} by taking a weighted average of all the individual entropy values.

\[Entropy Score = \alpha_{1}H_{1} + \alpha_{2}H_{2} + … + \alpha_{n}H_{n}\]

Where H\textsubscript{i} is the entropy computed for classifier \emph{i} and $\alpha$\textsubscript{i} is the weight assigned for the corresponding entropy. And $\sum_{i=1}^{n}\alpha$\textsubscript{i} = 1.

We can quantitatively compare the average entropy scores after running data generation jobs using different versions of the data generator.

For example, consider the situation where we have multiple versions of a conversation simulator - versions \emph{A}, \emph{B} and \emph{C} with entropy values listed in Table~\ref{tab:entropy1}. We would compute the weighted average of entropy values using equal weights for both entropy values. In this case, simulator version \emph{C} will rank the highest on the diversity axis, because of its entropy score.

\section{Conclusion and Future Work}
In this paper, we introduce the challenges associated with generating training data for conversational recommender systems and proposes a multi-dimensional approach towards the evaluation of the quality of such datasets synthesized using LLMs. Since we are only beginning to leverage the capabilities of LLMs, we expect future work in this domain to introduce more elaborate evaluation strategies that would enable researchers to apply rigorous quality checks on datasets synthesized using generative models.

\nocite{DBLP:journals/corr/abs-2004-00646, DBLP:journals/corr/abs-1904-02792, DBLP:journals/corr/abs-2003-02245, https://doi.org/10.48550/arxiv.2202.06935, https://doi.org/10.48550/arxiv.2107.00061, RICH1979329, Peng_2017, 10.1145/584091.584093, DBLP:journals/corr/abs-2201-08239, DBLP:journals/corr/abs-2201-08239, https://doi.org/10.48550/arxiv.2204.02311}

\section*{Limitations}
The approaches discussed in this paper have several limitations that could be addressed by future work on evaluation frameworks. The evaluation framework proposed in this paper is centered around two dimensions that we have found useful towards setting up evaluation criteria for finetuning datasets for conversational recommenders, namely, realism and diversity. However, the proposed evaluation does not take into account the quality of recommendations, the performance of the agent in terms of identifying user-preferences, eliciting high-signal information from the user, etc. The work in this paper is focused mainly on evaluating the quality of natural language dialogue in the conversational recommender.

\section*{Ethics Statement}
This paper aims to initiate a discussion around responsible evaluation of datasets generated by harnessing the capabilities of generative language models. In this work, we bring forth the idea that it is important to carefully evaluate the quality of training datasets generated using LLMs so that we can ensure good performance on downstream applications that rely on that training data.
\section*{Acknowledgements}
We would like to acknowledge the help and support of all the members on our team for reviewing the paper and providing valuable feedback.

% Entries for the entire Anthology, followed by custom entries
\bibliography{main}
\bibliographystyle{acl_natbib}

%%%%%%%%%%%%%%%%%%%%%%%%%%%%%%%%%%%%%%%%%%%%%%%%%%%%%%%%%%%%

%%%%%%%%%%%%%%%%%%%%%%%%%%%%%%%%%%%%%%%%%%%%%%%%%%%%%%%%%%%%

\appendix

\end{document}